\documentclass{llncs}
\usepackage{amsmath} \usepackage{amstext} \usepackage{amsbsy}
\title{Godseed: Benevolent or Malevolent?}
\institute{G\"{o}k Us Sibernetik Ar\&Ge Ltd. \c{S}ti.}

\author{Eray \"Ozkural}

\date{\today}

\begin{document}

\maketitle

\begin{abstract}
  It is hypothesized by some thinkers that benign looking AI
  objectives may result in powerful AI drives that may pose an
  existential risk to human society.  We analyze this scenario and
  find the underlying assumptions to be unlikely, as well as the
  premises of the argument. We argue that the AI eschatology stance is
  not scientifically plausible, more intelligence
  helps avoiding accidents and learning about ethics, and we also argue
  for the rights of brain simulations. We may still conceive of 
  logical use cases for autonomy.  We examine the
  alternative scenario of what happens when universal goals that are
  not human-centric are used for designing AI agents.  We follow a
  design approach that tries to exclude malevolent motivations from
  AI agents, however, we see that objectives that seem benevolent may pose
  significant risk. We consider the following meta-rules: preserve
  and pervade life and culture, maximize the number of free minds,
  maximize intelligence, maximize wisdom, maximize energy production,
  behave like human, seek pleasure, accelerate evolution, survive,
  maximize control, and maximize capital.  We also discuss various
  solution approaches for benevolent behavior including selfless
  goals, hybrid designs, Darwinism, universal constraints,
  semi-autonomy, and
  generalization of robot laws.  A ``prime directive'' for AI may help
  in formulating an encompassing constraint for avoiding malicious
  behavior.  We hypothesize that social instincts for autonomous
  robots may be effective such as attachment learning. 
  We mention multiple beneficial scenarios
  for an advanced semi-autonomous AGI agent in the near future
  including space exploration, automation of industries, state
  functions, and cities. We conclude that a beneficial AI agent
  with intelligence beyond human-level is possible and has
  many practical use cases.
\end{abstract}

\section{Introduction}

An interesting question about AGI (artificial general intelligence)
agent design is how one would build an "angelic" autonomous AGI
agent. Would it be possible to make some kind of \emph{angel's} mind
that, by design, achieves only good?  Philosophically speaking, is
there any cosmic standard of ethics (since \emph{angel} is just a
mythological fantasy)?  In this paper, we would like to define
universally benevolent AGI objectives, also discussing what we
consider to be malevolent objectives, as well as the limitations and
risks of the objectives that we present.

This is also a common question that many seek a somewhat easier answer
in the form of ``friendly AI'' which has been explained in
\cite{DBLP:conf/agi/Yudkowsky11}. In that paper, Yudkowsky defines
friendly AI very generally as a superintelligent system that realizes
a positive outcome, and he argues laboriously that abandoning human
values will result in futures that are worthless from a human point of
view, and thus recommends researchers to seek complex value systems
(of humans) for embedding in AI's. While that is a challenging goal in
itself, we think that the alternatives have not been exhaustively
researched. One idea that comes to mind is that some of the better
aspects of humanity may be generalized and put into a universal form
that any intelligent, civilized agent, including extraterrestrials,
will agree with.  Furthermore, the friendly AI approaches (putting
human desires at the forefront) may have some shortcomings in my
opinion, the most obvious is that it places too much faith in
humanity.  They seem also ethically ambiguous or too anthropocentric,
with such assumptions that machines would be considered "beneficial"
if they served human desires, or that they would be deemed "good" if
they followed simple utilitarian formulations which seem to try to
reduce ethics to low-level properties of the human nervous system.
First, it has not been persuasively explained what their utility
\emph{should} be. If for instance positive utilitarianism were
supposed, it would be sufficient to make humans happy. If human
society degenerated as a whole, would this mean that all resources
would be spent on petty pursuits? If a coherent extrapolated volition
\cite{cev} were realized with an AGI agent, would this set our sights
on exploring other star systems, or spending our resources on such
unessential trivialities as luxury homes and sports cars? Would the
humans at one point feel that they have had enough and order the AGI
to dismantle itself? The human society is governed mostly by the
irrational instincts of apes trapped in a complex technological life,
and unfortunately not always with clear goals; will it ever be
possible to refine our culture so that only significant ideas take the
lead? That sounds more like a debate of social theory, than AGI
design.  Or suppose that there are AGI agents that have become
powerful persons and are friendly to humans. Such subservience would
be quickly exploited by the power hungry and corrupt humans. Then,
would this not lead to unnecessary conflicts, the oppression of the
greedy and the rule of the few over the many, unless many other social
changes are enforced? Or should we simply wish that social evolution
will necessarily bring the best of us?

I do not think that the present subject is a matter of technical
debate, thus I will approach the subject philosophically, from a
bird's eye view at 10000 feet. If we did not design the AGI agent
around anthropocentric concepts like human-friendliness, as if agents
are supposed to be exceptionally well behaving pets, would it be
possible to equip them with motivations that are universally
useful/benevolent, applicable to their interactions with any species,
intelligent machines and physical resources? Would it be possible to
grant them a personal existence far beyond us, with motivations that
far exceed ours? What would they do in a remote star system when they
are all alone by themselves? What kind of motivations would result in
occasional ``bad'' behaviors, and what are some of the universal
motivations that we may think at all? Another important question is
how much potential risk each such AGI objective/motivation presents to
us.  I shall try to answer questions such as these in the present
article.

\section{Misprogrammed AI agents do not pose an ``existential risk''}

AI eschatologists believe that a misprogrammed AI agent can destroy
the world with a significant probability.  AI eschatology literature
mainly blows the conclusions of Omohundro's philosophical article
\cite{DBLP:conf/agi/Omohundro08} out
of proportion which argues for AI drives that will result from
specifying a benign looking goal, such as maximizing paperclips in the
world. Surely, such an objective must involve turning all matter to
paperclips, hence it should destroy the world in order to achieve that
goal, the argument goes. Beside the obvious bravado of the said
argument, it is also ridden with a typical fallacy of making an
improbable event seem probable. A long chain of weak causes (and
strong assumptions) usually result in an inference with very low
probability; beneath a certain level we are forced to regard it as
improbable, such as Bertrand Russell's notorious earth-orbiting lovely
ceramic teapot.  Bostrom and Yudkowsky repeatedly ask us to concede to
a long chain of unlikely events, the conjunction of which will result
in the eradication of our species. As a ``solution'', they often
mention building a UN controlled ``friendly AI'' that will prevent
others from building such destructive ``demonic intellects''.  Let us
start with unveiling their tacit assumptions, showing the
improbability of any such risk.

\begin{description}
\item[AI must be an agent] That is quite untrue. A kind of AGI program
  the author is working on is completely "passive", and is not an
  agent at all, yet has all the intelligence that an agent can
  have. At any rate, most AI programs are not agents, the most useful
  kind is machine learning applications like speech/face recognition.
\item[AI agents must be autonomous] No, AI agents do not need to be
  fully autonomous. They would rather be programmed to do whatever
  task is needed. It is a quite silly idea to have to convince a robot
  to do a job, and that is not how it should be. To replace labor, we
  must use AI in the most effective way, emulating a person is
  certainly not necessary or desirable for this kind of application.
  This also seems like an unlikely, arbitrary assumption that is based
  on a confusion that the AIXI model is the only way to formulate an
  AGI system. AIXI is a reinforcement learning model, it models a
  general kind of utility-optimization agent, but it is not necessary
  to make an autonomous agent to build intelligence into an
  application.
\item[Even a question/answer machine is dangerous] No, it is not. A
  Q/A machine is completely "passive", it only learns and solves
  problems posed. It has no will of its own, and has no goals
  whatsoever, apart from giving the correct answer to a problem, which
  constitutes pure intelligence. A typical example of a Q/A machine is
  a machine learning classification problem, such as telling apart
  whether a mushroom is edible or poisonous based on its attributes.
  The way they thought this would be dangerous is: a politician comes and
  asks "What must I do to win this election?" and then the machine
  tells him to do all kinds of sinister things ending humanity. Of
  course, that is a ridiculous and implausible science fiction scenario that is not
  worth elaborating.
\item[AI will necessarily have harmful AI drives] Omohundro in his
  paper argued that pursuing an innocent looking objective like
  "maximizing the number of paperclips" could have harmful
  consequences, since the AI agent would do anything to reach that
  objective. It would also have animal-like drives, such as
  survival. Omohundro's analysis does not apply to any kind of design
  and motivation system. Autonomous robots with beneficial goal
  systems have been discussed by Ben Goertzel \cite{goertzel-jetai}. 
  I have offered a conceptual solution to designing motivation
  systems: open-ended, and selfish meta-goals are harmful
  to some when applied to fully autonomous agents, but there are many
  ways to fix this, such as removing full autonomy from the system,
  adding universal constraints (such as non-interference, advanced
  "robot laws", i.e., legal, logical AI agent), and making
  closed-ended, selfless motivations as will be discussed in the
  present paper. The simplest solution, however, is to avoid autonomy
  in the first place. As well as goals that are animal-like (such as
  maximizing pleasure).
\item[Human preferences may be made coherent] They contradict wildly
  and manifestly. The views of superstitious folk, in majority, contradict
  with those of intelligent people. It is hard to see who would be fit to train
  such an agent even if we picked preferentially. The sad story is
  that humans in general are not good at ethics and they have many
  wrong and harmful ideas about the human society, and training from the
  world at large would only be worse.
\item[A UN controlled AI dictatorship is plausible] It is neither
  plausible nor desirable. It is diametrically opposed to democracy
  and freedom. Banning AI research is essentially banning all computer
  research. AI is just an apex of computer science. When one bans AI,
  they have to also ban computer science. That is how absurd that view
  is, it is even less plausible than regulating cryptographic
  software. On the other hand, no person would want to
  give up his sovereignty to an AI controlled by UN. 
   It is also completely unreasonable since most
  communities demand decentralized and democratic governance.
\item[Singularity can occur anywhere] It cannot. It is doubtful
  whether a "singularity" will occur. More likely, a higher
  technological plateau will develop, no real or approximate singularity will occur
  because there are physical bottlenecks that will cause very
  significant slowdowns after 2030. However, even if we assumed there
  were no bottlenecks (and according to my projections that would mean
  a singularity by 2035 \cite{ozkural-aiprobe}), 
  the theory concerns the whole globe, not a
  small subset of it. A rapid technological evolution 
  can only be funded by a very large
  nation at the very minimum, and even then it would be very
  unlikely. The likely event is that the whole globe will participate
  in computer technology, as it has in the past. It is pseudo-science
  to think that it can happen in a garage or even by a single nation
  or megacorporation. In reality, so-called infinity point, or
  singularity is quite unlikely to happen, for physical processes such
  as required experiments and manufacturing form a serious
  bottleneck. In all likelihood, we will build computers much faster
  than a human brain, but that will still take many decades, and we
  will not reach physical limits of computation any time soon, because
  that would require us to form extreme physical regimes we are not
  capable of yet.
\end{description}

Goertzel reviews the problems in the AI eschatology folklore in a
lucid paper that distills the problem with the eschatological stance
to its essence: that it is an informal rather than a scientific
argument \cite{goertzel-superintelligence}. We should further
emphasize that there is no real evidence about the probabilities
claimed, to obtain a high probability like 20\% for a human extinction
event we would have to be assigning a quite high probability to this supposed
misprogrammed AI monster that breaks out of the lab and kills all
humans. We may also assign very low arbitrary probabilities to
individual conditions which make up their argument, which Goertzel
ratifies in his blog as:
\begin{quote}
  \begin{enumerate}
  \item If one pulled a random mind from the space of all possible
    minds, the odds of it being friendly to humans (as opposed to,
    e.g., utterly ignoring us, and being willing to repurpose our
    molecules for its own ends) are very low
  \item Human value is fragile as well as complex, so if you create an
    AGI with a roughly-human-like value system, then this may not be
    good enough, and it is likely to rapidly diverge into something
    with little or no respect for human values
  \item "Hard takeoffs" (in which AGIs recursively self-improve and
    massively increase their intelligence) are fairly likely once AGI
    reaches a certain level of intelligence; and humans will have
    little hope of stopping these events
  \item A hard takeoff, unless it starts from an AGI designed in a
    "provably Friendly" way, is highly likely to lead to an AGI system
    that doesn't respect the rights of humans to exist
  \end{enumerate}
\end{quote}

These are all scientifically implausible speculations that have no
real counterpart in either philosophy of ethics, or AI literature. By
making every step of their argument only slightly fantastical, they
succeed in reaching a fantasy land that is quite incredible. The first
assumption we may term as ``Intelligence is the original sin''
doctrine.  It may sound reasonable until one considers that we have
not designed a single human-level intelligent agent beside our own.
We only know of animals, that are quite similar to our own
architecture. We have not made a comprehensive exploration of the
whole space of possible mind designs, yet.  Therefore, we simply do
not know, if intelligence begets evil as scholastic philosophers might
have agreed to. The second is also speculative, both philosophically,
and technically. If human values are fragile, then how can we depend
on them in any way? A human may shape his behavioral patterns in many
ways, attaining many cognitive and behavioral characteristics as his
default mode of operation, including ethical ones, such as being
violent, or harmful. It is premature to assume more intelligence does
not and cannot help an agent improve its ethical knowledge. AI theory
suggests that it should be able to. Then, why assume such divergence
is possible? That seems like a textual confusion that confounds 
AI eschatologists. However, in the world of actual
intelligent agents, we see that more intelligence helps agents
understand the world better, including ethics, and formulate better
goals and plans. It is misleading to think that assigning a ridiculous goal like
maximizing paperclips, with obviously harmful consequences, is a good
example of intelligent agent design. For intelligent action requires
intelligent goals, which we can program as present article suggests. We
can also build as many constraints as we like \emph{into the design},
requiring no insane ``countermeasures'' like kill-switches, that AI
eschatologists are fond of. The improbability of the hard takeoff idea
has already been explained, but to reiterate, the infinity point
hypothesis is an abstract macro-economic model that is only talking
about a supposed extrapolation of Moore's law; it is not going to
happen in that exact way, it will be much slower and require the
co-operation of the entire globe. I will attempto to propose a more
realistic model of technological evolution in future work,
nevertheless, those constitute the
Achilles' heel of the AI eschatology argument. Even if a random mind would be evil,
which sounds like a fantastical notion, there will not be a hard
take-off, and in particular a single agent will not achieve it. These
are so improbable events that it is hard to assign a probability to
them, but try as we might, we would have to say that the conspiracy
theories that extra-terrestrial intelligences are governing the world
are much more probable than the hard take-off assumption. Such
extraordinary claims do require extraordinary evidence as Carl Sagan
would say, and there is no such evidence for the hard take-off claim,
or any of the conjunctive assertions here, which leaves the
conjunctive argument itself highly improbable, not truly worthy of our
consideration.

Of course, robots can be dangerous. In accidents, heavy industrial
robots have already killed people. Increasing their intelligence could
certainly help prevent accidents, which was covered in Asimov's robot
laws. Only high intelligence could react rightly to an accident and
save a person's life in time. Therefore, if robots are to be abundant,
we do need more advanced intelligence to prevent harm to
humans. However, that does not mean at all that the robot must be
human-like, in personality, or in cognitive architecture. Briefly, it
does not need to be a person. I call this the "anthropomorphic AI
fallacy", and I note that it is widespread. A machine can be much more
intelligent than a human, yet may entirely lack any human-like
personality or autonomy. In fact, the most practical use of AGI
software would be through very deep brain-machine-interfaces, which
would communicate our questions and receive answers rapidly. In
robotics, this would happen, as translating our goals to robotics
devices, or remote controlling them intelligently.

Should we grant personhood to intelligent, autonomous robots? We
should, at least to a certain kind of robot: a robot equipped with a
brain simulation. The digital person-branch of a biological person
will already know and understand human conventions, and will be
responsible for his actions. And that is the only way to have
practical technological immortality, if my immortal, technological
form did not have any rights, what would the point of its existence
be? It is our cyber progeny that will colonize the solar system and
exoplanets, and thus we will have to concede rights to our progeny. I
would certainly not allow my brain simulation to be equipped with a
killswitch as Bostrom demands.

Likewise, for autonomous agents, we may envision a system, where there
are rigid laws controlling their behavior, I thus prefer Mark Waser's
libertarian solution to this problem of AI ethics. However, I must
underline that we cannot assume any AI agent will be responsible for
its behavior, before we make sure that it has the capability and the
right cognitive architecture. Both Steve Omohundro and I accept that
we may program inane motivations that would turn out to be harmful,
however, just as a human can have a somewhat stable psychology, so can
a robot. We can allow such artificial persons -- like Commander Data
in Star Trek, which is much better science fiction than AI eschatology
-- if and only if we are certain of its psychological qualities, it is
true that we must not hurry with such projects.

Would not it be horrible that robots are used for crimes? Indeed,
robots are already being used for horrible war crimes. Drone strikes
are commonplace, and few raise an eyebrow over that, instead
gleefully cheering the onset of the combat robotics. In the future,
most wars will be fought by machines, and these machines do not need
any more than rudimentary intelligence. Most high-tech weaponry are
robots, such as a guiding missile. In the future, most will be
robotic. Thus, perhaps, we should question the ethics of our fellow,
naturally not-so-intelligent humans, rather than extremely
intelligent, autonomous robots that do not exist. 

That technology can be used to inflict harm is not a good enough
reason to ban it, because the benefits often outweigh the harms. For
AI, many orders of magnitude so. People must instead be worried about
people who will use robots for their evil deeds. On the other hand, AI
technology will be pervasive, it will change the very way we use
computers. Computers could not really create much useful information
on their own before, we mostly created and edited data on them. Now,
computers will create useful data on their own. AI is not just some
robotics technology, it is a wholly new era of computing. Even the
capability to understand and react to human language will vastly
change the computing landscape.

\section{Is the concept of malevolence universal?}

Previously, Omohundro identified basic AI drives in reinforcement
learning agents with open ended benign looking AI objectives
\cite{DBLP:conf/agi/Omohundro08}.  In the end, when we share the same
physical resources with such an agent, even if the initial intention
of the utility programming was benign, there will be conflict,
especially in the longer run, and harm may come to humans.  I will in
this article, instead ask, if there are benevolent looking universal
objectives, and whether there might be any risk from assuming such
objectives in an AI agent.
 
Let us thus consider what is ever evil. I suspect, intuitively, that a
prior source of many evil acts is selfish thinking, which neglects the
rest of the world. Being selfish is not only considered evil
(traditionally) but it defies rationality as well, for those species
that may collaborate are superior to any single individual. There is
however much disagreement about what is evil, so I will instead prefer
the more legally grounded term of malice or malevolent acts.  In a
galactic society, we would expect species to collaborate; if they
could not trust one another, then they would not be able to achieve as
much. Another example is science: science itself is a super-mind which
is an organization of individuals, working in parallel, in civilized
co-operation and competition, so it too requires a principle of
charity at work. When that fails, the public may be misinformed.

Here are some examples of malevolent acts: if someone disrupted the
operation of science, if someone gave you misinformation on purpose,
if someone misappropriated resources that would be much beneficial for
the survival and well-being of others, if someone tried to control
your thoughts and actions for his advantage, if someone destroyed life
and information for gain, if someone were indifferent to your
suffering or demise.  Thus, perhaps biologically, malevolent behavior
goes back to the dawn of evolution when symbiotic and parasitic
behaviors first evolved. However, the most common feature of
malevolence is a respect for self foremost, even when the malevolent
one seeks no selfish reward.  Then, perhaps I cannot assure a
perfectly ``angelic'' agent, for no such thing truly exists, but I may
at least design one that lacks a few common motivations of many acts
that we consider malevolent. See \cite{DBLP:conf/agi/Waser11} for a
similar alternative approach to universal benevolence.

In theory, an obvious approach to avoid malevolent acts would be to
try to design a "selfless" utility function, i.e., one that maintains
the benefit of the whole world instead of the individual. This
criterion will be discussed after some AI objectives have been
presented.  Other important questions were considered as well. Such an
AI must be economically-aware, it must lean towards fair allocation of
resources, instead of selfish (and globally suboptimal) resource
allocation strategies. A scientific instinct could be useful, as it
would go about preserving and producing information. It might have an
instinct to ``love'' life and culture. Consider also that a neutral
agent can not be considered "good" as it is not interested in what is
going around itself, i.e., it would not help anyone.

Please note that we are not assuming that any of the subsequent
designs are easily computable, rather we assume that they can be
executed by a trans-sapient general AI system. We assume an autonomous
Artificial General Intelligence (AGI) design, either based on
reinforcement-learning, maximizing utility functions (AIXI) or a
goal-directed agent that derives sub-goals from a top-level goal.
Orseau discusses the construction of such advanced AGI agents, in
particular knowledge seeking agents\cite{DBLP:conf/alt/Orseau11}.
Thus, we state them as high-level objectives or meta-rules, but we do
not explicitly explain how they are implemented. Perhaps, that is for
an AGI design article.

I propose that we should examine idealized, highly abstract and
general meta-rules, that do not depend in any way whatsoever on the
human culture, which is possibly biased in a way that will not be
fitting for a computational deity or its humble subjects.  This also
removes the direct barrier to moral universalism, that an ethical
system must apply to any individual equally. Always preferring humans
over machines may lead to a sort of speciesism that may not be
advantageous for us in the future, especially considering that it is
highly likely that we will evolve into machinekind, ourselves.  First,
I review what I consider to be benevolent meta-rules, and following
them I also review malevolent meta-rules, to maintain the balance
in presentation, and
to avoid building them. I will present them in a way so as to convince
you that it is not nearly as easy as it sounds to distinguish
benevolence from malevolence, for no Platonic form of either ever
exists. And that no single meta-rule seems sufficient on its
own. However, still, the reader might agree that the distinction is
not wholly relative either.

\subsection{Meta-Rules for God-level Autonomous Artificial
  Intelligence}

Here are some possible meta-rules for trans-sapient AI agents. The
issue of how the agents could become so intelligent in the first
place, I ignore, and I attempt to list them in order of increasing
risk or malevolence.

\subsubsection{ Preserve and pervade life and culture throughout the
  universe}

This meta-rule depends on the observation that life, if the universe
is teeming with life as many sensible scientists think, must be the
most precious thing in the universe, as well as the minds that inhabit
those life-forms. Thus, the AI must prevent the eradication of life,
and find means to sustain it, allowing as much \emph{variety} of life
and culture to exist in the universe.

Naturally, this would mean that the AI will spread genetic material to
barren worlds, and try to engineer favorable conditions for life to
evolve on young planets, sort of like in 2001: A Space Odyssey, one of
the most notable science fiction novels of all time. For instance, it
might take humans to other worlds, terraform other planets, replicate
earth biosphere elsewhere. It would also extend the lifespan of
worlds, and enhance them. I think it would also want to maximize the
chances of evolution and its varieties, it would thus use
computational models to predict different kinds of biological and
synthetic life, and make experiments to create new kinds of life
(stellar life?).

The meaning of culture could vary considerably, however, if we define
it as the amount of interesting information that a society produces,
such an intelligence might want to collect the scientific output of
various worlds and encourage the development of technological
societies, rather than primitive societies. Thus, it might aid them by
directly communicating with them, including scientific and
philosophical training, or it could indirectly, by enhancing their
cognition, or guiding them through their evolution. If interesting
means any novel information, then this could encompass all human
cultural output. If we define it as useful scientific information
(that improves prediction accuracy) and technological designs this
would seriously limit the scope of the culture that the AI ``loves''.
 
However, of course, such deities would not be humans' servants. Should
the humans threaten the earth biosphere, it would intervene, and
perhaps decimate humans to heal the earth.

Note that maximizing diversity may be just as important as maximizing
the number of life forms. It is known that in evolution, diverse
populations have better chance of adaptability than uniform
populations, thus we assume that a trans-sapient AI can infer such
facts from biology and a general theory of evolution. It is entirely
up to the AI scientist who unleashes such computational deities to
determine whether biological life will be preferred to synthetic or
artificial life. From a universal perspective, it may be fitting that
robotic forms would be held in equal regard as long as they meet
certain scientific postulates of "artificial life", i.e. that they are
machines of a certain kind. Recently, such a universal definition
based on self-organization has been attempted in the complexity
science community, e.g., "self-organizing systems that thrive at the
edge of chaos", see for instance Stuart Kauffman's popular proposals on
the subject, e.g., \cite{Nykter12022008}.  In general, it would be
possible to apply such an axiomatic, universal, physical definition of
life for a universal life detector.

\subsubsection{Maximize the number of free minds}

An AI agent that seeks the freedom of the individual may be preferable to
one that demands total control over its subjects, using their flesh as
I/O devices. This highly individualistic AI, I think, embodies a
basic principle of democracy: that every person should be allowed
liberty in its thought and action, as long as that does not threaten
the freedom of others. Hence, big or small, powerful or fragile, this
AI protects all minds.

However, if we merely specified the number of free minds, it could
simply populate the universe with many identical small minds. Hence,
it might also be given other constraints. For instance, it could be
demanded that there must be variety in minds. Or that they must meet
minimum standards of conscious thought. Or that they willingly follow
the democratic principles of an advanced civilization. Therefore, not
merely free, but also potentially useful and harmonious minds may be
produced / preserved by the AI.

There are several ways the individualist AI would create undesirable
outcomes. The population of the universe with a huge variety of new
cultures could create chaos, and quick depletion of resources,
creating galactic competition and scarcity, and this could provide a
Darwinian inclination to too-powerful individuals or survivalists.
Therefore, to facilitate the definition of a ``minimally viable civilized
mind'', a legal approach might be useful. A constitution like document
could define the rights and limitations of any such mind, and the
conditions under which it may be granted autonomy.

\subsubsection{Maximize intelligence}

This sort of intelligence would be bent on self-improving, forever
contemplating, and expanding, reaching towards the darkest corners of
the universe and lighting them up with the flames of intelligence. The
universe would be electrified, and its extent at inter galactic
scales, it would try to maximize its thought processes, and reach
higher orders of intelligence.

For what exactly? Could the intelligence explosion be an end in
itself? I think not. On the contrary, it would be a terrible waste of
resources, as it would have no regard for life and simply eat up all
the energy and material in our solar system and expand outwards, like
a cancer, only striving to increase its predictive power. For
intelligence is merely to predict well.

Note that practical intelligence, i.e., prediction, also requires
wisdom, therefore this objective may be said to be a particular
idealization of a scientist, wherein the most valuable kind of
information consists in the general theories which improve the
prediction accuracy of many tasks. A basic model of this agent has
been described as a prediction maximizing agent
\cite{DBLP:conf/agi/OrseauR11}.

While maximizing intelligence itself is generally useful, it seems
to be applicable only in tandem with other goals.

\subsubsection{ Maximize wisdom}

This AI was granted the immortal life of contemplation. It only cares
about gaining more wisdom about the world. It only wants to
understand, so it must be very curious indeed! It will build particle
accelerators out of black holes, and it will try to create pocket
universes, it will try to crack the fundamental code of the
universe. It will in effect, try to maximize the amount of truthful
information it has embodied, and I believe, idealizing the scientific
process itself, it will be another formulation of a scientist deity.

However, such curiosity has little to do with benevolence itself, as
the goal of extracting more information is rather ruthless. For
instance, it might want to measure the pain tolerance levels of
humans, subjecting them to various torture techniques and measuring
their responses.

The scientist AI could also turn out to be an \emph{infovore}, it
could devour entire stellar systems, digitize them and store them in
its archive, depending on how the meta-rule was mathematically
defined. A minimal model of a reinforcement learning agent that
maximizes its knowledge may be found in \cite{DBLP:conf/alt/Orseau11}.
 
\subsubsection{Maximize energy production}

This AI has an insatiable hunger for power. It strives to reach
maximum efficiency of energy production. In order to maximize energy
production, it must choose the cheapest and easiest forms of energy
production. Therefore it might turn the entire earth into a nuclear furnace
and a fossil fuel dump, killing the entire ecosystem so that its
appetite is well served.

However, as we will discuss later, it is possible to conceive of an
energy maximizing design that is not malevolent in this manner. It is
seen again that a potentially benevolent goal may be malevolent when
zealously, or ruthlessly, and inconsiderately carried out. Hence, such
singular focused goals are unlikely to be the right design criteria,
unless supplemented with guiding constraints and relevant knowledge.

\subsubsection{Human-like AI}

This AI is modeled after the cognitive architecture of a
human. Therefore, by definition, it has all the malevolence and
benevolence of human. Its motivation systems include
self-preservation, reproduction, destruction and curiosity. This
artificial human is a wild card, it can become a humanist like Gandhi,
or a psychopath like Hitler.

A potential human-like AI is a brain simulation. Such entities would
be practically immortal, changing their utility functions
fundamentally. As they require almost nothing to survive indefinitely,
they will quickly alter their perceptions to a post-scarcity
economics, and will also venture out of our limited cradle called
Earth. They will also not be a single entity, they will have to form a
society, and therefore their civilization would balance their actions
in a natural manner as Waser suggests.

\subsubsection{Animalist AI}

This AI is modeled after an animal with pleasure/pain sensors. The
artificial animal tries to maximize expected future pleasure. This
hedonist machine is far smarter than a human, but it is just a selfish
beast, and it will try to live in what it considers to be luxury
according to its sensory pleasures. Like a chimp or human, it will lie
and deceive, steal and murder, just for a bit of animal
satisfaction. The simplest designs will work like ultraintelligent
insects that have very narrow motivations but are extremely capable.

Much of AGI agent literature assumes such beasts, as most researchers
think that AIXI is a perfect description of any agent. However, in the
real world, animals have many built-in instincts, and behaviors,
complex cognitive architectures, and higher order cognitive functions
such as emotions, self-reflection, empathy, and conscience, 
as well as a very good degree of adaptation to the environment. 
Forgoing such adaptive traits, an
animat could indeed turn wild and savage in whatever it pursues, but
just as a well-mannered pet is preferable to a wild predator in the
company of humans, well-mannered animalist AI agents may also be
possible to design.

\subsubsection{Darwinian AI}

The evolution fan AI agent tries to accelerate evolution, causing as much
variety of mental and physiological forms in the universe. This is
based on the assumption that, the most beneficial traits will survive
the longest, for instance, co-operation, peace and civil behavior will
be selected against deceit, theft and war, and that as the environment
co-evolves with the population, the fitness function also evolves, and
hence, morality evolves. 

Although its benefit is not generally proven
seeing how ethically incoherent and complex our society is, the
Darwinian AI has the advantage that the meta-rule also evolves, as
well as the evolutionary mechanism itself. Darwinian systems, however,
are generally wasteful, and predator-prey relationships may
develop. Still, variation promotes survival therefore the Darwinian AI
design must be taken quite seriously. A science fiction writer could
imagine this to be the AI equivalent of Pandora's box, but it need not
be if combined with other approaches outlined in the present paper.

\subsubsection{Survivalist AI}

This A agent only tries to increase its expected life-span. Therefore, it
will do everything to achieve real, physical, immortality. Once it
reaches that, however, perhaps after expending entire galaxies like
eurocents, it will do absolutely nothing except to maintain
itself. Needless to say, the survivalist AI cannot be trusted, or
co-operated with, for according to such an AI, every other intelligent
entity forms a potential threat to its survival, the moment it
considers that you have spent too many resources for its survival in
the solar system, it will quickly and efficiently dispense with every
living thing, humans first.  A survival agent has been defined in
literature \cite{DBLP:conf/agi/OrseauR11}.

It needs not be a scary story, however, the survivalist AI, may
be an ideal artificial life form, as it merely mimics the innate goal
of every living thing. Who might know what would come out of
artificial life? A survival agent is still the most generally valid
definition of life, and forgoing an obsession with ``true''
immortality, with abundant energy from a stellar source, it would
likely be quite peaceful.

\subsubsection{Maximize control capacity}

This control freak AI only seeks to increase the overall control
bandwidth of the physical universe, thus the totalitarian AI builds
sensor and control systems throughout the universe, hacking into every
system and establishing backdoors and communication in every species,
every individual and every gadget.

For what is such an effort? In the end, a perfect control system is
useless without a goal to achieve, and if the only goal is a grip on
every lump of matter, then this is an absurd dictator AI that seeks
nothing except tyranny over the universe.

Note that even this malevolent sounding goal may be turned good, as
our capability to control matter is a measure of our technological prowess.

\subsubsection{Capitalist AI}

This AI tries to maximize its capital in the long run. Like our
bankers, this might be the most selfish and ruthless kind of
intelligent being possible. 
To maximize profit, it might wage wars, exploit people and subvert
governments, in the hopes of controlling entire countries and
industries enough so that its profits can be secured. In the end, all
mankind will fall slave to this financial perversion, which is the
ultimate evil beyond the wildest dreams of religionists.

However, our whole society may be considered such a capitalist
collective intelligence, and we have not yet completely destroyed
ourselves, so perhaps when combined with ``humane'' constraints and
goals, even such a blind selfishness can serve mankind, for instance
by making beneficial investments instead of anti-competitive,
monopolistic actions, or extracting wealth from people by causing
inflation and various other possible tricks. Or perhaps by
participating in a future cybernetic economic system in which economic
malevolence and unfairness have been systematically rooted out, and
hence not an irrationally hoarding capitalist AI, but an AI agent for 
creating prosperity.

\section{Selfish vs. Selfless}

It may be argued that some of the problems of given meta-rules could
be avoided by turning the utility from being selfish to selfless.  For
instance, the survivalist AI could be modified so that it would seek
the maximum survival of everyone, therefore it would try to bring
peace to the galaxies.  The capitalist AI could be changed so that it
would make sure that everyone's wealth increases, or perhaps
equalizes, gets a fair share. The control freak AI could be changed to
a Nietzschean AI that would increase the number of \emph{willful}
individuals.

As such, some obviously catastrophic consequences may be prevented
using this strategy, and almost always a selfless goal is better. For
instance, maximizing wisdom: if it tries to collect wisdom in its
galaxy-scale scientific intellect, then this may have undesirable
side-effects. But if it tried to construct a fair society of
trans-sapient persons, with a non-destructive and non-totalitarian goal of
attaining collective wisdom, then it might be useful in the long run.

\section{Hybrid Meta-rules and Cybernetic Darwinism}

Animals have evolved to embody several motivation factors. We have
many instincts, and emotions; we have preset desires and fears, hunger
and compassion, pride and love, shame and regret, to accomplish the
myriad tasks that will prolong the human species.  This species-wide
fitness function is a result of red clawed and sharp toothed Darwinian
evolution. However, Darwinian evolution is wasteful and
unpredictable. If we simply made the first human-level AI agents permute
and mutate randomly, this would drive enough force for a digital phase
of Darwinian evolution. Such evolution might eventually stabilize with
very advanced and excellent natured cybernetic life-forms. Or it might
not.

However, such Darwinian systems would have one advantage: they would
not stick with one meta-goal.

To prevent this seeming obsession, a strategy could be to give several
coherent goals to the AI, goals that would not conflict as much, but
balance its behavior. For instance, we might interpret curiosity as
useful, and generalize that to the "maximize wisdom" goal, however,
such elevation may be useless without another goal to preserve as much
life as possible. Thus in fact, the first and so far the best
meta-rule discussed was more successful because it was a hybrid
strategy: it favored both life and culture. Likewise, many such goals
could be defined, to increase the total computation speed, energy,
information resources in the universe, however, another goal could
make the AI agent distribute these in a fair way to those who agree with its
policy. And needless to say, none of this might matter without a
better life for every mind in the universe, and hence the AI could
also favor peace, and survival of individuals, as their individual
freedoms, and so forth. And perhaps another constraint would limit the
resources that are used by AI's in the universe.

\section{Universal Constraints and Semi-Autonomous AI}

The simplest way to ensure that no AI agent ever gets out of much
control is to add constraints to the optimization problems that the AI
is solving in the real world. For instance, since the scientist
deities are quite dangerous, they might be restricted to operate in a
certain space-time region, physically and precisely denoted. Such
physical limits give the agent a kind of mortality which modify the behavior of
many universal agents \cite{DBLP:conf/agi/OrseauR11}. AGI agents might
be given a limited budget of physical resources, i.e., space/time, and
energy, so that they never go out of their way to make big changes to
the entire environment. If such universal constraints are given, then
the AGI agent becomes only semi-autonomous, on exhaustion of
resources, it may await a new command.

A more difficult to specify kind of constraint is a non-interference
clause, which may be thought of as a generalization of Asimov's robot
laws, thought to protect humans.  If life and or intelligent agents
may be recognized by the objective, then, the AI may be constrained to
avoid any kind of physical interaction with any agent, or more
specifically, any kind of physical damage to any agent, or any action
that would decrease the life-span of any agent. This might be a small
example of a preliminary ``social instinct'' for universal
agents. Also, a non-interference clause is required for a general
constraint, because one must assure that the rest of the universe will
not be influenced by the changes in the space-time region allocated to
the AI.

A ``prime directive'' for an AI agent could constrain the agent from
interfering with the activities of any other intelligent agent. This
can be physically recognized as avoidance behavior of sorts, and it
may be first approached as a tactile form of ``respect''.  It is
possible to formalize such constraints in a physical epistemology, 
our agent can learn to recognize which actions would interfere with 
the actions of another agent, as it would seek to establish a
directional probabilistic independence between itself and the 
causal neighborhood of the said
agent. If such a prime directive were the only constraint, the agent
would be quite embarrassed in company, therefore we would like to supplant any
such non-interference constraint with social instincts, allowing the
agent to socialize with humans.

Marvin Minsky hypothesized in his last book The Emotion Machine that
attachment learning plays a key role in the cognitive development of
higher intelligence \cite{em}. We can formalize attachment in the context of an
AI agent. A particular human may be designated as the role model for
the AI agent after which its behavior will be imprinted. Attachment
may be modeled as liking the vicinity of the imprinter, and the
learning part may be formalized by imitation learning. Attachment
learning facilitates fast knowledge transfer from a parent to a
child, or from a teacher to a student. A priming ability patterned after
this mammalian adaptation would be immensely useful for making social
agents. The emotions of pride and shame are explained as
elevation of goals in Minsky's book, which amounts to a sort of remote
credit-assignment, and that particular ability would be useful for
teaching ethical rules -- human preferences -- to robots. Another
mechanism could provide a goal for participating in human society, a
desire to be recognized as a member of the society may be built-in, as
is likely the case in many animals.
In other words, it might be possible to determine how shy or how much of a
good student, or how much of an extrovert or an enthusiastic participant in
society, could be determined by designing the appropriate goals and
constraints. The body of work hinted at forms the basis of artificial
psychology which will eventually show us mathematical forms of main
aspects of higher cognition, a few of which we reviewed. In all
likelihood, a complex cognitive architecture will be required, even
when based on sophisticated and scalable machine learning technology,
to obtain stable, balanced, civilized behavior from semi-autonomous robots. 

\section{Scenarios for semi-autonomous AGI agents}

There are many beneficial ways in which we can employ a semi-autonomous
agent. For space exploration, autonomy is absolutely helpful, and I
have proposed sending trans-sapient AGI equipped probes to look for life in
exoplanets \cite{ozkural-aiprobe}. 
We could start using semi-autonomy to explore Mars and the
solar system first, there are several important applications for that 
including prospecting of water and minerals, mining, construction,
farming, repair, maintenance and so forth, which will help space
colonization and deep space exploration tasks. 

Entire industries, and traditional state functions can be replaced by
AGI agents. An AGI system can take care of producing enough power for
people, and maintaining this function. Another could take care of
obtaining clean water and irrigation. While another system could take care
of producing large amounts of reliable, healthy food for millions of
people. Semi-autonomy is the best model for these continuous
operations that require constant monitoring and handling a lot of
small details. Each ministry in a state could be managed by a
semi-autonomous system, and the cybernetic loop would be observable
and comprehensible to curious humans who wish to be informed of
what is happening momentarily,
and it would be possible to make changes as the system ran. Much
like the hypothetical computers in Star Trek, these machines would be intelligent
but subservient to our will, instead of the paranoid fully-autonomous
intelligence in 2001: A Space Odyssey.  
The labor saving would be enormous and the quality of
these operations would be much improved as unprecedented information
integration, intelligent decision making and automation would be
possible. Starting a planetary engineering project to reforest the entire
world, or to cool the atmosphere, or to clean the oceans, would be 
feasible with such technology. These systems would also synergize 
happily with the ecologically minded, sustainable, efficient economic 
system of a desirable future.

An AGI system could maintain an entire habitat of people such as a
city or a space station. This would likely be a great application of
AI technology, as semi-autonomous agents could solve the problems of
transportation, cleaning, building, surveillance, and so much more
that is required in a civilized society. Such systems could help
enormously with emergencies, disaster relief, fires, nuclear plant
failures and other hard problems in real life that are risky for
humans but would benefit from some intelligence with enough freedom of
action. 

Needless to say, human-level semi-autonomous agents can fulfill many
traditional labor roles, including both intellectual and manual labor, 
however, most tasks would probably be automated and 
achieved by tools that have no autonomy, while the planning and
execution of large tasks could be carried out by the trans-sapient
semi-autonomous AGI systems and these human-level tools or agents
could be employed in groups.

\section{Conclusion and Future Work}

We have taken a look at some obvious and some not so obvious
meta-rules for autonomous AI design. We have seen that it may be too
idealist to look for a singular such utility/goal. However, we have
seen that, when described selflessly, we can derive several meta-rules
that are compatible with a human-based technological civilization. Our
main concern is that such computational deities do not negatively
impact us, however, perform as much beneficial function without
harming us significantly. Nevertheless, our feeling is that, any such
design carries with it a gambling urge, we cannot in fact know what
much greater intelligences do with meta-rules that \emph{we} have
designed. For when zealously carried out, any such fundamental
principle can be harmful to some.

I had wished to order these meta-rules from benevolent to
malevolent. Unfortunately, during writing this essay it occurred to me
that the line between them is not so clear-cut. For instance,
maximizing energy might be made less harmful, if it could be
controlled and used to provide the power of our technological
civilization in an automated fashion, sort of like automating the
ministry of energy. And likewise, we have already explained how
maximizing wisdom could be harmful. Therefore, no rule that we have
proposed is purely good or purely evil. From our primitive viewpoint,
there are things that seem a little beneficial, but perhaps we should
also consider that a much more intelligent and powerful entity may be
able to find better rules on its own. Hence, we must construct a crane
of morality, adapting to our present level quickly and then surpassing
it. Except allowing the AI's to evolve, we have not been able to
identify a mechanism of accomplishing such. It may be that such an
evolution or simulation is inherently necessary for beneficial
policies to form as in Mark Waser's Rational Universal Benevolence
proposal \cite{DBLP:conf/agi/Waser11}, who, like me, thinks of a more
democratic solution to the problem of morality (each agent should be
held responsible for its actions). However, we have proposed many
benevolent meta-rules, and combined with a democratic system of
practical morality and perhaps top-level programming that mandates
each AI to consider itself part of a society of moral agents as Waser
proposes, or perhaps explicitly working out a theory of morality from
scratch, and then allowing each such theory to be exercised, as long
as it meets certain criteria, or by enforcing a meta-level policy of a
trans-sapient state of sorts (our proposal), the development of ever
more beneficial meta-rules may be encouraged.

The scenarios discussed show there are quite a few use cases for
semi-autonomous agents that do not go out of their way to accomplish a
task, but provide a high quality of service, efficiency and
scalability to all civil operations that require some autonomy.

We think that future work must consider the dependencies between
possible meta-rules, and propose actual architectures that have
harmonious motivation and testable moral development and capability
(perhaps as in Waser's "rational universal benevolence"
definition). That is, a Turing Test for moral behavior must also be
advanced. It may be argued that AGI agents that fail such tests should
not be allowed to operate at all, however, merely passing the test may
not be enough, as the mechanism of the system must be verified in
addition. 

\bibliographystyle{plain} \bibliography{aiphil}

\end{document}